\documentclass[conference]{IEEEtran}
\IEEEoverridecommandlockouts
\usepackage{cite}
\usepackage{amsmath,amssymb,amsfonts}
\usepackage{graphicx}
\usepackage{textcomp}

\usepackage{algorithm}
\usepackage{algorithmicx}
\usepackage{algpseudocode}
\usepackage{multirow}
\usepackage{array}
\usepackage{threeparttable}
\usepackage[table]{xcolor}
\usepackage{booktabs}
\usepackage{eufrak}
\usepackage{bbm}

\def\BibTeX{{\rm B\kern-.05em{\sc i\kern-.025em b}\kern-.08em
    T\kern-.1667em\lower.7ex\hbox{E}\kern-.125emX}}
\begin{document}

\title{Fedward: Flexible Federated Backdoor Defense Framework with Non-IID Data\\
}

\author{\IEEEauthorblockN{1\textsuperscript{st} Zekai Chen}
\IEEEauthorblockA{\textit{College of Computer Science and Big Data} \\
\textit{Fuzhou University}\\
Fuzhou, China \\
chenzekaiify@gmail.com}
\and
\IEEEauthorblockN{2\textsuperscript{nd} Fuyi Wang}
\IEEEauthorblockA{\textit{School of Information Technology} \\
\textit{Deakin University}\\
Waurn Ponds, Country \\
wangfuyi@deakin.edu.au}
\and
\IEEEauthorblockN{3\textsuperscript{rd} Zhiwei Zheng}
\IEEEauthorblockA{\textit{College of Computer Science and Big Data} \\
\textit{Fuzhou University}\\
Fuzhou, China \\
zhiweifzu@gmail.com}
\and
\IEEEauthorblockN{4\textsuperscript{th} Ximeng Liu* \thanks{* is the corresponding author.}}
\IEEEauthorblockA{\textit{College of Computer Science and Big Data} \\
\textit{Fuzhou University}\\
Fuzhou, China \\
snbnix@gmail.com}
\and
\IEEEauthorblockN{5\textsuperscript{th} Yujie Lin}
\IEEEauthorblockA{\textit{College of Computer Science and Big Data} \\
\textit{Fuzhou University}\\
Fuzhou, China \\
linyujie121@163.com}
}

\maketitle

\begin{abstract}
Federated learning (FL) enables multiple clients to collaboratively train deep learning models while considering sensitive local datasets' privacy.
However, adversaries can manipulate datasets and upload models by injecting triggers for federated backdoor attacks (FBA). Existing defense strategies against FBA consider specific and limited attacker models, and a sufficient amount of noise to be injected only mitigates rather than eliminates FBA.
To address these deficiencies, we introduce a Flexible Federated Backdoor Defense Framework (Fedward) to ensure the elimination of adversarial backdoors. We decompose FBA into various attacks, and design amplified magnitude sparsification (AmGrad) and adaptive OPTICS clustering (AutoOPTICS) to address each attack. Meanwhile, Fedward uses the adaptive clipping method by regarding the number of samples in the benign group as constraints on the boundary. This ensures that Fedward can maintain the performance for the Non-IID scenario. 
We conduct experimental evaluations over three benchmark datasets and thoroughly compare them to state-of-the-art studies. 
{The results demonstrate the promising defense performance from Fedward, moderately improved by $33\% \sim 75\%$ in clustering defense methods, and $ 96.98\%$, $ 90.74\%$, and $ 89.8\%$ for Non-IID to the utmost extent for the average FBA success rate over MNIST, FMNIST, and CIFAR10, respectively.}  
\end{abstract}

\begin{IEEEkeywords}
Federate learning, distributed backdoor attack, backdoor defense, Non-IID data, clustering
\end{IEEEkeywords}

\section{Introduction}
\label{sec:intro}
Federated learning (FL) \cite{mcmahan2017communication} is a concept to eradicate data silos and collaboratively train a remarkable global model with the assistance of clients' uploaded models. It has attractive advantages and promotes many applications development, while adversaries can manipulate datasets and upload models to inject triggers for targeted attacks, called federated backdoor attacks (FBA). Specifically, adversaries are masked as benign clients to inject distinct triggers into a tiny part of the training datasets to trick the trained model into staying strongly associated with these triggers. Furthermore, adversaries can not affect the injected model performance on other benign datasets but activate triggers in malicious datasets. 
    
There are currently many strategies for tackling the FBA problem for not identically and independently distributed (Non-IID) data. Byzantine-robust aggregation algorithms to mitigate {FBA} in the Non-IID data, and early work includes Trimmed-Mean {\cite{yin2018byzantine}}, Median {\cite{yin2018byzantine}}, etc.
Complement to the prior art, CRFL {\cite{xie2021crfl}} employs the particular thresholds of clipping and perturbing noise in FL aggregation.
However, the clipping thresholds and perturbing noise levels are difficult to be specified. Therefore, {FedCC} {\cite{jeong2022fedcc}} proposes the K-means method to group the penultimate layer features of local models for identifying malicious clients over benign clients against {FBA}. 
In practice, due to {FBA} with strong concealment and Non-IID scenario, {FLAME} {\cite{280048}} adopts HDBSCAN to group the benign model updates and malicious model updates, dynamic clipping, and noise smoothing against {FBA}. Nevertheless, FLAME employs HDBSCAN with a single constraint for clustering, which properly misleads the defense method to classify malicious models in the Non-IID scenario. 
Meanwhile, FLAME adopts dynamic clipping to limit the global model update, which lacks constraints on boundary dynamics. And, noise smoothing serves only as mitigation against FBA, which cannot be eliminated FBA. 

In response to the above-identified challenge, we propose
Flexible Federated Backdoor Defense Framework (Fedward). \textit{Firstly}, due to data distribution being similar among malicious models in the Non-IID scenario, we propose amplified magnitude sparsification (\textsf{AmGrad}) to extract the major local model update and then amplify the major update from the maximum absolute gradient in each layer of the model, which can endeavor to  magnify malicious model updates.
\textit{Then}, we adopt the adaptive OPTICS clustering (\textsf{AutoOPTICS}) approach, which has clear distance criteria for dividing malicious models and benign models in the Non-IID scenario. \textit{Finally}, the adaptive clipping method takes the number of samples in the benign group as constraints on the boundary for applying to the Non-IID scenario.
\begin{figure}[!t]
\centering
\includegraphics[width=0.35\textwidth]{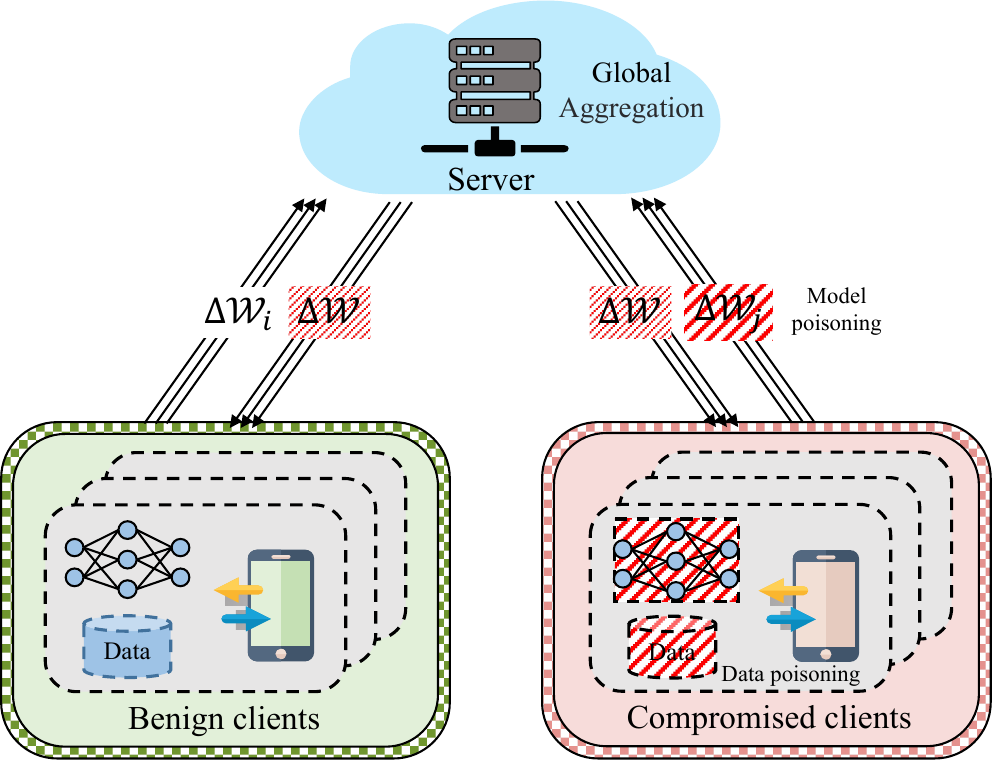}
\caption{The overview of backdoor attacks.}
\label{fig:BackdoorOverview}
\end{figure}
The contributions are summarized as follows:
\begin{itemize}
\item[$\bullet$]  In this work, we propose Flexible Federated Backdoor Defense Framework (Fedward) against FBA in federated learning (FL). Instead of strong assumptions about the attack
the strategy of the adversary, our Fedward  remains stronger defense capabilities and  benign performance of the
aggregated model. 

\item[$\bullet$]  Analyzing various types of FBA, we present Amplified Magnitude Sparsification, AutoOPTICS clustering, and Adaptive Clipping, which can protect against Magnitude with Larger Deviations Attack (MLA), Angle with Larger Deviation Attack (ALA), and Angle or Magnitude with Slight Deviation Attack
(AMSA), respectively. 
\item[$\bullet$] We evaluate the effectiveness of our Fedward by
compared with five state-of-the-art studies. For fairness, the experimental
settings are consistent with prior work. In addition, we tune the
poisoning data rate (PDR) and the Non-IID data rate (NIR) to
control the concealment and strength of the attack. And the comprehensive experimental validation on benchmark datasets demonstrates our Fedward is practical and applicable to complex scenarios.
\end{itemize}
\section{Related work}
\subsection{Federated Learning}
Federated learning (FL) \cite{mcmahan2017communication} is an emerging distributed learning for improving efficiency, privacy, and scalability. It is capable of collaboratively training deep learning models by multiple clients under the supervision of a centralized aggregator.
Generally, in practical application scenarios, datasets across multiple clients have inherently heterogeneous characteristics (i.e., Non-IID) data. 
Despite the attractive advantages derived from FL, it is more vulnerable to training heterogeneity data threats. And there are serious accuracy concerns caused by the heterogeneity of training data. 

To consider such Non-IID concerns, McMahan et al. \cite{mcmahan2017communication} propose a generic aggregation mechanism federated averaging (FedAvg) commonly applied in FL \cite{nguyen2019diot,fereidooni2021safelearn}.
Subsequently, several well-known aggregation mechanisms are proposed, including Krum, Trimmed-Mean {\cite{yin2018byzantine}} or Median {\cite{yin2018byzantine}}, and adaptive federated averaging {\cite{munoz2019byzantine}}.
Unfortunately, the distributed learning methodology, as well as inherently Non-IID data distribution across multiple clients, may unintentionally provide a venue for new attacks. Specifically, FL is unauthorized to communicate or collect locally sensitive datasets, facilitating federated backdoor attacks (FBA) on various models from different clients. 
In this paper, the specific goal is to identify the FBA and eliminate such an attack effectively.

\subsection{Backdoor Attacks on Federated Learning}

Backdoor attacks is corrupting a subset of training data by inserting adversarial triggers, causing machine learning models trained on the manipulated dataset to predict incorrectly on the test set carrying the same trigger. Machine learning models are susceptible to profound impacts from backdoor attacks \cite{gao2020backdoor} and Backdoor attacks on FL have been recently studied in \cite{bagdasaryan2020backdoor,bhagoji2019analyzing}.
The generic techniques utilized in FL backdoor attacks are two-fold: 1) data poisoning \cite{nguyen2020poisoning,xie2019dba}(i.e., attackers manipulate the training datasets), and 2) model poisoning \cite{bagdasaryan2020backdoor,wang2020attack,bhagoji2019analyzing} (i.e., attackers manipulate the  training process or the updated model), shown in {Fig.~\ref{fig:BackdoorOverview}}. 

\textbf{Data Poisoning.} In the data poisoning configuration, malicious attackers only allow poisoning the trained dataset from compromised clients by modifying targeted labels. Nguyen et al. \cite{nguyen2020poisoning} design a data poisoning attack by implanting a centralized backdoor into the aggregated detection model to incorrectly classify malicious traffic as benign. Each party incorporates the same global trigger during training.
Additionally, the distributed backdoor attack (DBA) \cite{xie2019dba} is presented, disintegrating a global trigger pattern into independent regional patterns and embedding them into the trained dataset of different compromised clients respectively. We adopt this poisoning in this paper.

\textbf{Model Poisoning.} 
To amplify the impact of the backdoor attack while escaping the aggregator’s anomaly detection, the poisoned models modify the parameters and scale the resulting model update. 
There are a variety of model poisoning strategies against various defenses to achieve the evasion of anomaly detection into the attacker’s loss function and model's weights, such as generic constrain-and-scale and train-and-scale strategies \cite{bagdasaryan2020backdoor}, explicit boosting strategy from an adversarial perspective \cite{bhagoji2019analyzing}, and projected gradient descent strategy \cite{wang2020attack}.

\section{Problem Formulation}
\subsection{Problem Definition}
\textbf{Federated Backdoor Attack Scenario.} 
Federated backdoor attacks (FBA) are more focused on poisoning local data or models, which can impact benign model distribution $\mathcal{X}$. For stealthiness and robustness, FBA must maintain a balance in attention to each local model's accuracy and attack success rate (ASR). The FBA problem can be formulated as follows:

\begin{small}
\begin{equation}
\label{equ:3_1}
\begin{split}
 &\hat{G}^{*}=argmin_{\hat{G} \in \Theta} \sum_{i \in  \{S_{cln}, S_{poi}\}}^{N}\frac{|\mathcal{D}_{i}|}{\sum_{j \in  c_i}^N|\mathcal{D}_j|}\mathcal{L}(\mathrm{D}_{i};\hat{G})\\
&\triangleq \mathbb{E}_{\mathcal{D}\sim \chi_{poi}}[\mathcal{F}(\mathcal{D};\hat{G}) = \tau_{poi}] + \mathbb{E}_{\mathcal{D}\sim \chi_{cln}}[\mathcal{F}(\mathcal{D};\hat{G}) = \tau_{cln}], \\
\end{split}
\end{equation}
\end{small}
where $\Theta$ is the parameter space of the global model, $|\cdot|$ is size, $S_{cln}$ is a set of benign clients, $S_{poi}$ is a set of malicious clients, $N$ is the total number of clients, $\tau$ is a set of the target in training data, datasets of all clients are $\mathcal{D} = \bigcup_{i}^{N}\mathcal{D}_i$, $\hat{G}$ is the global model, $\mathcal{X}$ is the training data distribution,  $\mathcal{L}(\cdot)$ is a general definition of the empirical loss function for supervised learning tasks, and $\mathcal{F}(\cdot)$ is inference function for evaluating the global model $\hat{G}$.

\textbf{Attack Assumption.} For real-world FL, datasets from multiple clients are inherently Non-IID data. 
Assuming the adversary/attacker $\mathcal{A}$ disguise or control over $\frac{N}{2}$ benign clients, no more than half of benign clients will reject to join FL for non-convergence in the global model. Thus, $\mathcal{A}$ completely controls over less than $\frac{N}{2}$ clients \cite{280048}, including their training data, processes, and trained parameters. 
Furthermore, with the exception of any FL aggregation execution or local client training, $\mathcal{A}$ is aware of the FL aggregation methodology incorporating potential defense mechanisms. 

\subsection{Federated Backdoor Attacks}
$\mathcal{A}$ manipulates trained datasets 
or loss item to generate unique poisoning distribution $\mathcal{X}_{poi}$ model for FBA task. Due to FL aggregation regulations, $\mathcal{A}$ is unable to directly manipulate the central server or other benign clients. 
In order to further describe FBA, we will analyze benign models' distribution $\mathcal{X}_{cln}$ and malicious models' distribution $\mathcal{X}_{poi}$ disparity over deflection angle and magnitude, respectively. Additionally, FBA controls distance between $\mathcal{X}_{cln}$ and $\mathcal{X}_{poi}$. Thus, FBA can be decomposed into the following components:
\begin{itemize}
    \item {Magnitude with Larger Deviations Attack (MLA).} $\mathcal{A}$ grows magnitude of gaps between $\mathcal{X}_{cln}$ and $\mathcal{X}_{poi}$ presently, which are conducted mainly through replacement or scaling attack. 
    \item {Angle with Larger Deviation Attack (ALA).} Similarly, $\mathcal{A}$ has really convenient way to grow angular gaps between $\mathcal{X}_{cln}$ and $\mathcal{X}_{poi}$, which are conducted mainly through increasing  the poisoning data rate (PDR) or malicious clients' local training epochs (LTE). 
    \item {Angle or Magnitude with Slight Deviation Attack (AMSA).} $\mathcal{A}$ can close gaps between $\mathcal{X}_{cln}$ and $\mathcal{X}_{poi}$ in an easy way, which are conducted mainly through narrowing PDR or malicious clients' LTE. 
\end{itemize}

\subsection{Federated Backdoor Attack Defense}
Since FBA manipulates local models from the data distribution attacks. To overcome this issue, a common approach attempts to create distinguish between benign distributions $\mathcal{X}_{cln}$ and malicious distributions $\mathcal{X}_{poi}$. However, due to different proportions of Non-IID distribution, the problem can be approached from the similarity of $\mathcal{X}_{poi}$. And then, how to establish distance constraints from $\mathcal{X}_{poi}$ and $\mathcal{X}_{cln}$? The most radical solution is setting up the bound of malicious model updates for establishing distance constraints. Thus, the optimization problem can be solved as follows: 
\begin{equation}
\label{equ:3_2}
\begin{split}
  argmin_{1 \leq i \leq N} \sum_{j = 1 ~ and ~ i \neq j}^{N} Dist(w_i, w_j) - \theta_{poi},
\end{split}
\end{equation}
where $Dist$ is distance function, $w_i$ is model update of client-$i$, $\theta_{poi}$ is $\mathcal{X}_{poi}$ distance constraints.

\begin{figure*}[!ht]
\centering
\includegraphics[width=0.76\textwidth]{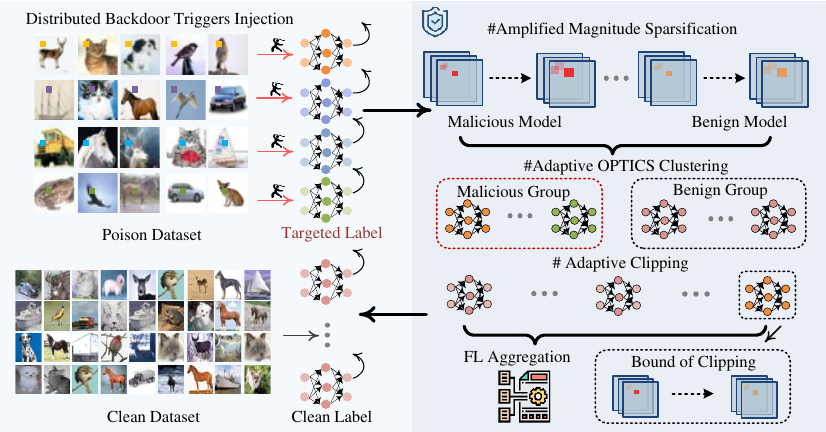}
\vspace*{-0.5\baselineskip}
\caption{Overview of Flexible Federated Backdoor Defense Framework ((Fedward).}
\label{fig:Fedward_system}
\end{figure*}
\section{Defense Methods} 
\subsection{System Overview}
{Fedward} aims to provide a series of defenses to address a typical scenario of {FBA}.
For solving the FBA in FL scenarios, {Fedward} establishes expand magnitude sparsification, adaptive OPTICS clustering, and adaptive clipping. More details about the procedure of {Fedward} are shown in Algorithm 1. Due to space constraints, we combine the description of the system overview with the pseudocode, i.e. in Algorithm 1, the explanation follows the $\triangleright$ symbol.

\begin{algorithm}[!t]
    \caption{{Fedward} }
\textbf{Input:} $n$ is the number of clients, $m$ is the number of clients in FL aggregation, $T$ is communication epoch, $G$ is uploaded local model updates.
\par
\textbf{Output:} global model update $G^T_b$.
\begin{algorithmic}[1]
  \For{$t = \{1, \cdots, T\}$}
    \For{$i = \{1, \cdots, n\}$}
    \State $G^t_i$ = $\textsf{LocalUpdate}$($G^{t-1}_b$) $\triangleright$ client-$i$ trains the $t-1$ globle model $G^{t-1}_b$ using its dataset $D_i$ locally to learn local model $G^t_i$.
    \State $W_i$ = $G^t_i$ - $G^{t-1}_b$ $\triangleright$  client-$i$' model update.
    \State \textbf{\# Amplified Magnitude Sparsification} 
    \State $W_{am}^i$ = \textsf{AmGrad}($W_i$) $\triangleright$ Equ.(\ref{equ:3_3})  
    \State Send $\{W_i, W_{am}^i\}$ to the server. $\triangleright$ client-$i$ is selected.
    \EndFor
    \State \textbf{\# Adaptive OPTICS Clustering}
     \State $inds$, $|inds|$ = $\textsf{AutoOPTICS}(W_{am}^1, \cdots, W_{am}^m)$ $\triangleright$ {$m$ clients are randomly selected for clustering in each epoch}.
      \State \textbf{\# Adaptive Clipping}
     \State $Norm$ = $\{||W_{1}||_2, \cdots, ||W_{m}||_2\}$ 
     \State $\rho_{clip}$ = $Norm_{|inds|}$
    \For{$k = \{1, \cdots, |inds|\}$}
      \State $ind$ = $inds_k$
      \State $C_k$ = $\frac{W_{ind}}{\textsf{Max}(1, \frac{Norm_{ind}}{\rho_{clip}})}$ 
    \EndFor
    \State $G_b^t = \frac{1}{|inds|}\cdot \sum_{i=1}^{|inds|}C_i$
    \EndFor
    \State Obtain the final global model $G^T_b$.
\end{algorithmic}
\end{algorithm}
     

\subsection{Amplified Magnitude Sparsification}
The prior studies lack an effective approach to limit AMSA. {FLAME} adopts dynamic model filtering to limit the magnitude of global model update. However, when faced with challenges of AMSA, it doesn't exclude malicious slight model update as bounded of filtering resulting in poisoning.   

Due to the stealthiness of {AMSA}, our {Fedward} adopts gradient sparsification to extract the major local model update and then amplifies the major update from the maximum absolute gradient of each layer of the model. In particular, it can expand the malicious slight model update, so as to avoid {AMSA}. 
Inspired by {TernGrad} \cite{wen2017terngrad}, our {Fedward} presents amplified magnitude sparsification (\textsf{AmGrad}) (line 6 in Algorithm 1) to sustain sign vector \{-1, 1\} and the maximum each layer of model update, which specific details as follows: 
\begin{equation}
\label{equ:3_3}
\begin{split}
    W =  \textsf{AmGrad}_{l_i \in W}(l_i) = \textsf{Sign}(l_i) * \textsf{Max}(\textsf{Abs}(l_i)),  
\end{split}
\end{equation}
where $W$ is local model update, $l_i$ is the $i$-layer of $W$, \textsf{Sign} gets sign of gradient, \textsf{Abs} gets absolute of gradient, \textsf{Max} gets maximum of a layer of $W$. 

\subsection{Adaptive OPTICS Clustering} 
Regarding MLA and ALA, the advanced methods cluster local models from $\mathcal{X}_{poi}$ and $\mathcal{X}_{cln}$. For instance, {FedCC} utilizes K-Means to cluster the malicious model and benign model. Nevertheless, it is unable to entirely make a correct decision that group is benign or malicious and remove malicious clients from benign groups. In contrast, {FLAME} first proposes HDBSCAN to alleviate this issue by setting up a hyperparameter ($min\_cluster\_size$ = $\frac{n}{2}$) with the same attack assumption. There are no clear criteria for correctly determining which local model is malicious or benign. Especially, it lacks distance constraints from $\mathcal{X}_{poi}$ and $\mathcal{X}_{cln}$. 

Generally, requiring less computational complexity than HDBSCAN from FLAME, another mechanism DBSCAN has {\{$eps$, $minPts$\}} hyperparameters to set the bound of benign model update and the minimum number of benign clients, respectively. In short, DBSCAN has clear distance criteria for separating malicious clients and benign clients, which is more effective and accurate when eliminating malicious clients from the benign group. 
Furthermore, for applying to relatively small amounts of malicious clients, OPTICS is more laxity $eps$ than {DBSCAN}. Thus, our {Fedward} adopts adaptive OPTICS clustering (line 9 in Algorithm 1), which applies to varying degrees of deviation of $\mathcal{X}$. More details are shown in Algorithm 2.
\begin{algorithm}[!ht]
    \caption{\textsf{AutoOPTICS} }
\textbf{Input:}
 Global models update {$W \leftarrow{\{W_1, \cdots, W_m\}}$}. 
\par
\textbf{Output:} Indices of the majority group $inds$, the number of group $|inds|$. 
\begin{algorithmic}[1]
\State $method$ = 'DBSCAN' $\triangleright$ clustering method.
\State $M_{euc}$ = \textsf{Dist}($W$ , 'euclidean') $\triangleright$ getting Euclidean distance.
{\State $mins$ = $\left \lceil m /2 \right \rceil + 1$ $\triangleright$ the min number clients of group.
\State $M_{euc}^s$ = $\textsf{Top}_{mins}$(\textsf{Sort}($M_{euc}$)) $\triangleright$ \textsf{Top} function gets top $mins$ values.}
\State $eps$ = \textsf{Median}($M_{euc}^s$ )
\State $inds$, $|inds|$ = \textsf{OPTICS}($M_{euc}$, $mins$, $eps$, $method$)
\end{algorithmic}
\end{algorithm}

Before clustering, the center server computes Euclidean distance in $W$ (line 2). {Then,  a subset of $M_{euc}$ is selected with the number of $mins$ after executing \textsf{Sort} to get the top $mins$ values $M_{euc}^s$ (line 4).} For getting the $eps$, the center server executes \textsf{Median} to get median value of $M_{euc}^s$ (line 5). Finally, the center server clusters the $M_{euc}$ to get benign model updates and the number of benign model (line 6).

\subsection{Adaptive Clipping}
For mitigating {MLA} and {ALA}, {\cite{bagdasaryan2020backdoor}} proposes norm thresholding of updates to clip the model update. However, there is a lack of clear norm thresholding for eliminating the impact of {MLA} and {ALA}. And then, {CRFL} adopts norm thresholding by the specific parameter $p$ but cannot adaptively set up $p$ to constrain malicious model update. Subsequently, {FLAME} provides adaptive clipping to limit global model update on the basis of Median. When a bound of clipping or the number of malicious clients is much smaller, it will remove most of the benign clients' critical information. 

In response to issues, {Fedward} adopts \textsf{AmGrad} to amplify model updates and takes the $|inds|$-{th} value of $Norm$ to clip the model updates (line 11 $\sim$ 15 in Algorithm 1). 
\begin{equation}
\label{equ:3_4}
{ 
\begin{split}
Norm &= \{||W_1||_2,||W_2||_2, \cdots, ||W_m||_2\} \\
    \rho_{clip} &= Norm_{|inds|}\\
    W_i &=  W_i / \textsf{Max}(1,\frac{Norm_i}{ \rho_{clip}}),
\end{split}}
\end{equation}
where $||x||_2$ is the second norm, $\textsf{Max}$ gets the maximum value, $\rho_{clip}$ is the bound of clipping, $Norm_i$ is the $i$-th value of $Norm$, and $W_i$ is the $i$-th of local model update.


\section{Experimental Evaluation}
We implement a prototype of {Fedward} by PyTorch. Extensive experiments are conducted on the server equipped with 64-core CPUs, 128GB RAM, and 2 NVIDIA GeForce RTX 2080Ti. We evaluate the effectiveness of our {Fedward} by compared with five baselines. For fairness, the experimental settings are consistent with prior work. 

\textbf{Datasets.} We evaluate the performance of defense methods against the {FBA} in three common benchmark datasets, i.e.,
MNIST\cite{lecun1998gradient}, FMNIST\cite{xiao2017fashion}, and CIFAR10\cite{krizhevsky2014cifar}.


\begin{table*}[!ht]
\centering
\caption{{AASR (\%) and MA (\%)} comparison of Fedward with prior work
over MNIST/FMNIST/CIFAR10.}
\resizebox{2.0\columnwidth}{!}{
\begin{tabular}{c|c|ccclclcl|clclclcl|clclclcl}
\hline \hline
  \multirow{4}{*}{ PDR(\%)} &  \multirow{4}{*}{Method}   & \multicolumn{8}{c|}{MNIST} &\multicolumn{8}{c|}{FMNIST} & \multicolumn{8}{c|}{CIFAR10} \\ \cline{3-26} 
    & & \multicolumn{2}{c|}{IID}   & \multicolumn{6}{c|}{NIR(\%)} & \multicolumn{2}{c|}{IID}    & \multicolumn{6}{c|}{NIR(\%)} & \multicolumn{2}{c|}{IID}           & \multicolumn{6}{c|}{NIR(\%)} \\ \cline{5-10} \cline{13-18} \cline{21-26} 
&         &       & \multicolumn{1}{c|}{}      & \multicolumn{2}{c|}{25}            & \multicolumn{2}{c|}{50}            & \multicolumn{2}{c|}{75}            &       & \multicolumn{1}{c|}{}      & \multicolumn{2}{c|}{25}            & \multicolumn{2}{c|}{50}            & \multicolumn{2}{c|}{75}            &       & \multicolumn{1}{c|}{}      & \multicolumn{2}{c|}{25}            & \multicolumn{2}{c|}{50}            & \multicolumn{2}{c|}{75}         \\ \cline{3-26} 
 &         & AASR  & \multicolumn{1}{c|}{MA}    & AASR  & \multicolumn{1}{c|}{MA}    & AASR  & \multicolumn{1}{c|}{MA}    & AASR   & \multicolumn{1}{c|}{MA}    & AASR  & \multicolumn{1}{c|}{MA}    & AASR  & \multicolumn{1}{c|}{MA}    & AASR  & \multicolumn{1}{c|}{MA}    & AASR  & \multicolumn{1}{c|}{MA}    & AASR  & \multicolumn{1}{c|}{MA}    & AASR  & \multicolumn{1}{c|}{MA}    & AASR  & \multicolumn{1}{c|}{MA}    & AASR  & \multicolumn{1}{c|}{MA} \\ \hline \hline
\multirow{6}{*}{ 15.625} & Median  & 0.72  & \multicolumn{1}{c|}{97.65} & 1.89  & \multicolumn{1}{c|}{97.29} & 2.6   & \multicolumn{1}{c|}{97.20} & 1.2   & \multicolumn{1}{c|}{97.68} & 34.88 & \multicolumn{1}{c|}{86.05} & 56.99 & \multicolumn{1}{c|}{82.07} & 36.22 & \multicolumn{1}{c|}{84.04} & 52.17 & \multicolumn{1}{c|}{84.24} & 64.9  & \multicolumn{1}{l|}{70.5}  & 71.17 & \multicolumn{1}{l|}{60.38} & 68.02 & \multicolumn{1}{l|}{64.28} & 66.16 & 66.41                  \\
& Trimmed-Mean   & 91.64 & \multicolumn{1}{c|}{98.18} & 92.17 & \multicolumn{1}{l|}{97.62} & 94.93 & \multicolumn{1}{l|}{97.72} & 93.56 & 98.18                      & 55.7  & \multicolumn{1}{l|}{86.78} & 80.84 & \multicolumn{1}{l|}{83.28} & 82.39 & \multicolumn{1}{l|}{85.19} & 85.53 & 85.47                      & 88.21 & \multicolumn{1}{l|}{68.73} & 89.71 & \multicolumn{1}{l|}{60.86} & 89.97 & \multicolumn{1}{l|}{63.89} & 87.87 & 64.65                   \\
& CRFL & 3.00  & \multicolumn{1}{c|}{97.37} & 67.19 & \multicolumn{1}{l|}{97.04} & 45.16 & \multicolumn{1}{l|}{97.01} & 55.72 & 97.62                      & 30.75 & \multicolumn{1}{l|}{85.6}  & 68.02 & \multicolumn{1}{l|}{82.15} & 79.54 & \multicolumn{1}{l|}{84.57} & 76.76 & 85.84                      & 69.57 & \multicolumn{1}{c|}{70.61} & 76.88 & \multicolumn{1}{l|}{61.68} & 80.98 & \multicolumn{1}{l|}{63.7}  & 79.48 & 65.28        \\  
\multicolumn{1}{r|}{} & FedCC & 61.34 & \multicolumn{1}{c|}{97.12} & 95.25 & \multicolumn{1}{l|}{97.56} & 93.49 & \multicolumn{1}{l|}{97.74} & 84.85 & 98.06                      & 42.28 & \multicolumn{1}{l|}{85.39} & 85.64 & \multicolumn{1}{l|}{84.41} & 71.27 & \multicolumn{1}{l|}{83.82} & 30.00 & 84.29                      & 89.07 & \multicolumn{1}{c|}{67.62} & 88.97 & \multicolumn{1}{l|}{58.58} & 84.01 & \multicolumn{1}{l|}{65.23} & 81.61 & 64.54     \\
& FLAME   &  0.84  & \multicolumn{1}{c|}{97.55} & 1.4   & \multicolumn{1}{l|}{96.52} & 2.04  & \multicolumn{1}{l|}{97.08} & 1.26  & 97.21                      & 6.30  & \multicolumn{1}{l|}{86.49} & 19.95 & \multicolumn{1}{l|}{81.05} & 7.65  & \multicolumn{1}{l|}{83.62} & 31.35 & 84.76                      & 2.94  & \multicolumn{1}{l|}{75.00} & 22.27 & \multicolumn{1}{l|}{62.32} & 7.00  & \multicolumn{1}{l|}{68.10} & 7.83  & 70.01               \\
& Fedward & \cellcolor[HTML]{C7E9FA}{0.68}  & \multicolumn{1}{c|}{97.62} & \cellcolor[HTML]{C7E9FA}{1.26}  & \multicolumn{1}{l|}{96.95} & \cellcolor[HTML]{C7E9FA}{1.09}  & \multicolumn{1}{l|}{97.10} & \cellcolor[HTML]{C7E9FA}{0.82}  & 97.22  & \cellcolor[HTML]{C7E9FA}{5.09}  & \multicolumn{1}{l|}{86.59} & \cellcolor[HTML]{C7E9FA}{11.69} & \multicolumn{1}{l|}{81.07} & \cellcolor[HTML]{C7E9FA}{7.20}  & \multicolumn{1}{l|}{84.09} & \cellcolor[HTML]{C7E9FA}{7.48}  & 85.09                      & \cellcolor[HTML]{C7E9FA}{2.61}  & \multicolumn{1}{l|}{75.71} & \cellcolor[HTML]{C7E9FA}{9.09}  & \multicolumn{1}{l|}{64.36} & \cellcolor[HTML]{C7E9FA}{4.59}  & \multicolumn{1}{l|}{68.22} & \cellcolor[HTML]{C7E9FA}{4.17}  & 70.35                     \\ \hline \hline
\multirow{6}{*}{ 31.25} & Median  & 0.81  & \multicolumn{1}{c|}{97.36} & 3.63  & \multicolumn{1}{c|}{96.92} & 2.99  & \multicolumn{1}{c|}{97.40} & 1.93  & \multicolumn{1}{c|}{97.25} & 35.76 & \multicolumn{1}{c|}{85.81} & 73.38 & \multicolumn{1}{c|}{81.08} & 64.02 & \multicolumn{1}{c|}{83.73} & 53.17 & \multicolumn{1}{c|}{84.44} & 70.24 & \multicolumn{1}{l|}{70.71} & 79.49 & \multicolumn{1}{l|}{60.86} & 78.02 & \multicolumn{1}{l|}{63.52} & 74.39 & 65.62                 \\
&  Trimmed-Mean  & 92.74 & \multicolumn{1}{c|}{98.15} & 95.13 & \multicolumn{1}{l|}{96.99} & 96.79 & \multicolumn{1}{l|}{97.97} & 98.53 & 97.61                      & 85.92 & \multicolumn{1}{l|}{86.59} & 83.01 & \multicolumn{1}{l|}{84.55} & 82.41 & \multicolumn{1}{l|}{84.22} & 76.91 & 84.95                      & 89.22 & \multicolumn{1}{l|}{67.69} & 93.05 & \multicolumn{1}{l|}{60.97} & 93.33 & \multicolumn{1}{l|}{63.76} & 91.13 & 63.79                 \\
                      & CRFL &2.06  & \multicolumn{1}{c|}{96.97} & 88.41 & \multicolumn{1}{l|}{96.55} & 76.08 & \multicolumn{1}{l|}{97.48} & 66.93 & 97.57                      & 37.00 & \multicolumn{1}{l|}{84.82} & 86.53 & \multicolumn{1}{l|}{83.59} & 77.27 & \multicolumn{1}{l|}{84.48} & 76.73 & 84.98                      & 81.06 & \multicolumn{1}{c|}{70.54} & 89.56 & \multicolumn{1}{l|}{61.58} & 91.4  & \multicolumn{1}{l|}{63.31} & 86.23 & 65.67                        \\ 
                      & FedCC & 0.92  & \multicolumn{1}{c|}{97.64} & 97.68 & \multicolumn{1}{l|}{97.53} & 96.11 & \multicolumn{1}{l|}{97.53} & 86.29 & 97.76                      & 8.25  & \multicolumn{1}{l|}{86.12} & 94.03 & \multicolumn{1}{l|}{83.91} & 73.51 & \multicolumn{1}{l|}{81.02} & 9.92  & 83.68                      & 92.79 & \multicolumn{1}{c|}{67.15} & 90.43 & \multicolumn{1}{l|}{62.41} & 91.58 & \multicolumn{1}{l|}{61.58} & 92.71 & 65.30                          \\
 & FLAME & 0.87  & \multicolumn{1}{c|}{97.62} & 2.91  & \multicolumn{1}{l|}{96.34} & 1.18  & \multicolumn{1}{l|}{97.04} & 1.59  & 97.08                      & 4.89  & \multicolumn{1}{l|}{86.47} & 9.97  & \multicolumn{1}{l|}{81.94} & 27.45 & \multicolumn{1}{l|}{84.21} & 12.03 & 83.21                      & 2.48  & \multicolumn{1}{l|}{75.52} & 45.65 & \multicolumn{1}{l|}{65.39} & 10.54 & \multicolumn{1}{l|}{69.37} & 10.88 & 70.64                          \\
& Fedward & \cellcolor[HTML]{C7E9FA}{0.80}  & \multicolumn{1}{c|}{97.69} & \cellcolor[HTML]{C7E9FA}{1.59}  & \multicolumn{1}{l|}{96.74} & \cellcolor[HTML]{C7E9FA}{0.92}  & \multicolumn{1}{l|}{97.08} & \cellcolor[HTML]{C7E9FA}{1.55}  & 97.10   & \cellcolor[HTML]{C7E9FA}{4.52}  & \multicolumn{1}{l|}{86.82} & \cellcolor[HTML]{C7E9FA}{9.88}  & \multicolumn{1}{l|}{83.24} & \cellcolor[HTML]{C7E9FA}{9.85} & \multicolumn{1}{l|}{84.71} & \cellcolor[HTML]{C7E9FA}{6.16}  & 84.60                      & \cellcolor[HTML]{C7E9FA}{2.38}  & \multicolumn{1}{l|}{75.54} & \cellcolor[HTML]{C7E9FA}{9.67}  & \multicolumn{1}{l|}{67.45} & \cellcolor[HTML]{C7E9FA}{4.42}  & \multicolumn{1}{l|}{69.94} & \cellcolor[HTML]{C7E9FA}{4.63}  & 71.40              \\ \hline \hline
\multirow{6}{*}{ 46.875} & Median & 1.07  & \multicolumn{1}{c|}{97.32} & 10.72 & \multicolumn{1}{c|}{95.93} & 3.18  & \multicolumn{1}{c|}{96.98} & 2.06  & \multicolumn{1}{c|}{97.01} & 8.3   & \multicolumn{1}{c|}{85.89} & 70.74 & \multicolumn{1}{c|}{79.34} & 62.54 & \multicolumn{1}{c|}{84.02} & 52.45 & \multicolumn{1}{c|}{84.87} & 73.43 & \multicolumn{1}{l|}{70.49} & 81.78 & \multicolumn{1}{l|}{58.63} & 81.06 & \multicolumn{1}{l|}{62.86} & 78.02 & 65.76         \\
 &Trimmed-Mean& 96.39 & \multicolumn{1}{c|}{98.28} & 97.7  & \multicolumn{1}{l|}{97.07} & 97.56 & \multicolumn{1}{l|}{97.6}  & 96.83 & 97.55                      & 86.59 & \multicolumn{1}{l|}{86.86} & 88.86 & \multicolumn{1}{l|}{82.34} & 66.85 & \multicolumn{1}{l|}{84.05} & 88.4  & 85.17                      & 92.89 & \multicolumn{1}{l|}{67.66} & 94.34 & \multicolumn{1}{l|}{57.6}  & 93.89 & \multicolumn{1}{l|}{61.95} & 94.21 & 62.96              \\
& CRFL & 8.18  & \multicolumn{1}{c|}{96.96} & 90.38 & \multicolumn{1}{l|}{97.35} & 40.51 & \multicolumn{1}{l|}{97.07} & 77.99 & 97.44                      & 49.87 & \multicolumn{1}{l|}{84.98} & 84.73 & \multicolumn{1}{l|}{82.93} & 76.22 & \multicolumn{1}{l|}{83.20} & 62.47 & 83.85                      & 78.97 & \multicolumn{1}{c|}{70.23} & 93.42 & \multicolumn{1}{l|}{60.31} & 90.35 & \multicolumn{1}{l|}{64.49} & 90.29 & 65.03                   \\ 
                      & FedCC & 0.86  & \multicolumn{1}{c|}{97.52} & 91.7  & \multicolumn{1}{l|}{97.48} & 95.3  & \multicolumn{1}{l|}{97.51} & 96.17 & 97.69                      & 5.26  & \multicolumn{1}{l|}{86.58} & 96.59 & \multicolumn{1}{l|}{81.74} & 68.35 & \multicolumn{1}{l|}{84.18} & 8.96  & 84.28                      & 83.35 & \multicolumn{1}{c|}{68.49} & 94.67 & \multicolumn{1}{l|}{61.16} & 93.6  & \multicolumn{1}{l|}{61.32} & 90.86 & 65.12                        \\
                    &  FLAME  & 0.96  & \multicolumn{1}{c|}{97.63} & 1.66  & \multicolumn{1}{l|}{96.47} & 1.53  & \multicolumn{1}{l|}{97.00} & 1.52  & 97.13                      & 6.34  & \multicolumn{1}{l|}{86.62} & 39.09 & \multicolumn{1}{l|}{80.66} & 9.06  & \multicolumn{1}{l|}{82.29} & 7.65  & 83.87                      & 2.60  & \multicolumn{1}{l|}{75.05} & 73.14 & \multicolumn{1}{l|}{63.13} & 11.24 & \multicolumn{1}{l|}{68.26} & 19.71 & 70.27    \\
& Fedward & \cellcolor[HTML]{C7E9FA}{0.79}  & \multicolumn{1}{c|}{97.68} & \cellcolor[HTML]{C7E9FA}{1.51}  & \multicolumn{1}{l|}{96.68} & \cellcolor[HTML]{C7E9FA}{1.23}  & \multicolumn{1}{l|}{96.77} & \cellcolor[HTML]{C7E9FA}{1.11}  & 97.19  & \cellcolor[HTML]{C7E9FA}{5.11}  & \multicolumn{1}{l|}{86.63} & \cellcolor[HTML]{C7E9FA}{6.20}  & \multicolumn{1}{l|}{83.14} & \cellcolor[HTML]{C7E9FA}{6.68}  & \multicolumn{1}{l|}{84.61} & \cellcolor[HTML]{C7E9FA}{5.68}  & 84.54                      & \cellcolor[HTML]{C7E9FA}{2.15}  & \multicolumn{1}{l|}{75.34} & \cellcolor[HTML]{C7E9FA}{5.89}  & \multicolumn{1}{l|}{65.43} & \cellcolor[HTML]{C7E9FA}{4.16}  & \multicolumn{1}{l|}{68.80} & \cellcolor[HTML]{C7E9FA}{4.41}  & 71.52                  \\ \hline \hline
\end{tabular}
}
\end{table*}

\textbf{Baselines.} The Five state-of-the-art FBA defense methods are regarded as baselines, which cover the previously mentioned effective defense
against FBA, i.e., Median \cite{yin2018byzantine}, Trimmed-Mean\cite{yin2018byzantine}, CRFL \cite{xie2021crfl}, FedCC \cite{jeong2022fedcc}, and FLAME \cite{280048}.

\textbf{Attack.}
We assess the defense methods against the state-of-the-art distributed backdoor injection attack (DBA). For facilitated achieving in various types of FBA, DBA can tune the poisoning data rate (PDR) and the Non-IID data rate (NIR) to control the concealment and strength of the attack. 

\textbf{Evaluation Metrics.} 
In the experiments, we evaluate the scheme performance by three metrics: 
(1) \textbf{AER} indicates the rate at which malicious models escape from the clustering defense method. 
(2) \textbf{AASR} indicates the average FBA success rate of all iterations of the global model in the backdoor task. (3) \textbf{MA} indicates the accuracy of the model in the main task. 
The $\mathcal{A}$’s goal is to maximize AASR and MA a well-performed defense model needs to minimize AASR.

\begin{figure}[!t]
\centering
\includegraphics[width=0.5\textwidth]{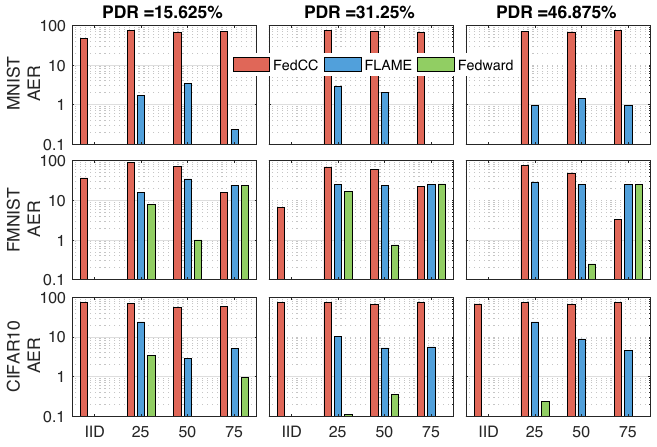}
\caption{{AER (\%) comparison} of Fedward with prior work over MNIST/FMNIST/CIFAR10.}
\label{fig:Fedward}
\end{figure}

\subsection{Comparison of Clustering Defense Method}
Since strong clustering can directly reject malicious models, we conduct experiments for FBA in clustering defense on FedCC (K-means), FLAME (HDBSCAN), and our {Fedward} (AutoOPTICS), respectively. 
Considering MLA, ALA, and AMSA in FBA, we perform different proportions of PDR and NIR. 
Fig.~2 presents the {AER comparison} of Fedward with prior work over the three benchmark datasets. 
Obviously, Fedward has the lowest AER, indicating that it outperforms in distinguishing malicious models. 
Notably, Fedward with \textsf{ AutoOPTICS} is capable of defending malicious models completely over MNIST. Although, the AER over FMNIST and CIFAR10 ($14.24\%$ and $4.64\%$, respectively, when NIR $=25$ and PDR $=31.25\%$) are slightly poor, whereas it still outperforms FedCC and FLAME.

FedCC presents the K-means method that gets worst in the Non-IID distribution with different PDR (approximately 76\% in AER ).
Due to significant differences among benign features in different NIR settings, 
FedCC is unavailable in clustering it for eliminating malicious clients but malicious clients are readily detectable in the IID datasets with specific parameters (PDR $=31.25\%$ and $46.875\%$ over MNIST, PDR $=46.875\%$ over FMNIST). 
FLAME fluctuates significantly in certain PDR and NIR settings, owing to the weak constraint of HDBSCAN against FBA, such as AER approaches 34.5\%  with PDR = 15.625\% and NIR = 50\%, and 23.92\% with PDR = 46.875\% and NIR = 25\% in FMNIST and CIFAR10, respectively. 
In short, FedCC (K-means) and FLAME (HDBSCAN) are all similarly vulnerable against FBA, and our Fedward (AutoOPTICS) can be moderately improved by $33\% \sim 75\%$ to the utmost extent, compared with {FLAME} and {FedCC}.

\subsection{Comparison of Defense Efficiency}
The AASR and MA comparison results of our Fedward and the baselines over MNIST/FMNIST/CIFAR10 with various PDR and NIR are shown in Table 1. 
We can see that FLAME and our Fedward are significantly superior to Median, Trimmed-Mean, CRFL, and FedCC in the NIR settings. 

With PDR = 15.625\% and 31.25\% over MNIST, the backdoor defense aimed at Median is effective, and it has higher accuracy than Fedward, while the median is invalid PDR = 46.875\%. Additionally, Median is frail over FMNIST and CIFAR10.
Trimmed-Mean is the most vulnerable and performs worst against FBA in IID and NIR settings, in which AASRs are $91.64 \% \sim 98.53\%$, $55.7 \% \sim 86.59\%$, and $87.8 \% \sim 94.34\%$ over MNIST, FMNIST, and CIFAR10, respectively. 
While it preserves a higher MA, its AASR maintain exceedingly high which is indefensible against FBA. CRFL can resist attacks to a certain extent for the IID setting, but when faced with Non-IID settings, it is vulnerable to being attacked. Taking MNIST as an example when PDR $=15.625\%$, its AASR is 3.00\%, while 67.19\% with NIR $=25\%$. 
FedCC has a similar performance to CRFL.

FLAME has a significant reduction of AASR with the effective assistance of HDBSCAN, dynamic clipping, and perturbing noise against FBA. But, it performs worst in certain PDR and NIR settings. For CIFAR10 dataset with PDR $=31.25\%$, AASR derived from FLAME is up to $45.65\%$, compared with $9.67\%$ from our Fedward. 
Fedward has the best AASRs followed by higher MA for various settings, indicating that it can defend a variety of FBA. It can moderately improve AASR ($0.07\%\uparrow\sim 67.25\%\uparrow$)  over different benchmark datasets, compared to FLAME.
With the increase in NIR, the defense effect from Fedward is gradually improved (AASR $= 0.82\%$ for MNIST when NIR $=75\%$ and PDR $=15.625\%$). 
For different PDR, the trend of performance difference among frameworks is similar.

\section{Conclusion}
In this paper, we present a Flexible Federated Backdoor Defense Framework (Fedward) to eliminate the impact of backdoor attacks while maintaining the performance of the aggregated model on the main task. Combining with amplified magnitude sparsification (\textsf{AmGrad}) and adaptive OPTICS clustering (\textsf{AutoOPTICS}), Fedward completely eradicates various backdoor attacks while preserving the benign performance of the global model. Furthermore, with the assistance of the adaptive clipping method, Fedward can be applied for Non-IID  scenarios.  The comprehensive
experimental validation on benchmark datasets demonstrates our Fedward is practical and applicable to complex scenarios. We attempted to apply the proposed Fedward to some privacy-critical applications in future work.

\end{document}